\documentclass[conference]{IEEEtran}
\IEEEoverridecommandlockouts

\usepackage{cite}
\usepackage{amsmath,amssymb,amsfonts}
\usepackage{algorithmic}
\usepackage{graphicx}
\usepackage{textcomp}
\usepackage{xcolor}
\def\BibTeX{{\rm B\kern-.05em{\sc i\kern-.025em b}\kern-.08em
    T\kern-.1667em\lower.7ex\hbox{E}\kern-.125emX}}

\usepackage[T1]{fontenc}
\usepackage{multirow}
\usepackage{tkz-graph}
\usepackage{url}
\newcommand{\varDataSetsNumber}{50}

\definecolor{blue(pigment)}{rgb}{0.2, 0.2, 0.6}
\definecolor{blond}{rgb}{0.98, 0.94, 0.75}
\definecolor{greenhtml}{rgb}{0.0, 0.5, 0.0}
\definecolor{dartmouthgreen}{rgb}{0.05, 0.5, 0.06}
\definecolor{icterine}{rgb}{0.99, 0.97, 0.37}
\definecolor{blizzardblue}{rgb}{0.67, 0.9, 0.93}
\definecolor{brilliantlavender}{rgb}{0.96, 0.73, 1.0}
\definecolor{carolinablue}{rgb}{0.6, 0.73, 0.89}
\definecolor{flamingopink}{rgb}{0.99, 0.56, 0.67}
\definecolor{lilac}{rgb}{0.78, 0.64, 0.78}
\definecolor{magicmint}{rgb}{0.67, 0.94, 0.82}
\definecolor{mossgreen}{rgb}{0.68, 0.87, 0.68}
\definecolor{mulberry}{rgb}{0.77, 0.29, 0.55}

\definecolor{midnightgreen(eaglegreen)}{rgb}{0.0, 0.29, 0.33}
\GraphInit[vstyle = Shade]
\tikzset{
	LabelStyle/.style = { rectangle, rounded corners, draw,
		minimum width = 2em, fill = blond!80,
		text = blue(pigment), font = \bfseries },
	VertexStyle/.append style = { inner sep=3pt,black,
		font = \Large\bfseries},
	EdgeStyle/.append style = {->, bend left} }

\newboolean{anonymous}
\setboolean{anonymous}{false}

\begin{document}

\title{Fuzzy Cognitive Maps and Hidden Markov Models: Comparative Analysis of Efficiency within the~Confines of the~Time Series Classification Task
\thanks{The~project was funded by POB Research Centre for Artificial Intelligence and Robotics of Warsaw University of Technology within the~Excellence Initiative Program - Research University (ID-UB).
}}

\ifthenelse{\boolean{anonymous}}
{
\author{\IEEEauthorblockN{First Name Middle Name Last Name}
\IEEEauthorblockA{\textit{Institution/Organization} \\
City, Country\\
\url{email@address}}
\and
\IEEEauthorblockN{First Name Middle Name Last Name}
\IEEEauthorblockA{\textit{Institution/Organization} \\
City, Country \\
\url{email@address}}}
}
{
\author{\IEEEauthorblockN{Jakub Micha\l ~Bilski}
\IEEEauthorblockA{\textit{Faculty of Mathematics and Information Science} \\
\textit{Warsaw University of Technology}\\
Warsaw, Poland}
\and
\IEEEauthorblockN{Agnieszka Jastrzebska}
\IEEEauthorblockA{\textit{Faculty of Mathematics and Information Science} \\
\textit{Warsaw University of Technology}\\
Warsaw, Poland \\
\url{A.Jastrzebska@mini.pw.edu.pl}}}
}
\maketitle

\begin{abstract}
Time series classification is one of the very popular machine learning tasks. In this paper, we explore the application of Hidden Markov Model (HMM) for time series classification. We distinguish between two modes of HMM application. The first, in which a single model is built for each class. The second, in which one HMM is built for each time series. We then transfer both approaches for classifier construction to the domain of Fuzzy Cognitive Maps. The identified four models, HMM NN (HMM, one per series), HMM 1C (HMM, one per class), FCM NN, and FCM 1C are then studied in a series of experiments. We compare the performance of different models and investigate the impact of their hyperparameters on the time series classification accuracy. The empirical evaluation shows a clear advantage of the one-model-per-series approach. The results show that the choice between HMM and FCM should be dataset-dependent. 
\end{abstract}


\section{Introduction}

The goal of time series classification is to assign correct class labels to time series based on information collected from a~training set of samples. 
This task is related to standard pattern classification, but the difference is that the attributes are ordered and this ordering matters.
There are many different real world domains in which we need to classify time series.

Importantly, there is a~huge qualitative difference between the different time series data sets that need to be classified. 
For example, in some domains we collect long time series, but the characteristic feature that is the basis for classification appears only once, at an unknown point in time. 
In other applications, there is a~pattern whose frequency of occurrence determines class belongingness. 
Still in other domains, we can directly compare levels of phenomena at consecutive points in time.
Undoubtedly, the wealth of diverse  time series data sets that need to be processed in the real world requires new methods and exploration of existing approaches.

The literature of temporal data modelling offers a number of interesting models that can be described as state-based. Such models, whose roots can be seen in the automata theory, are represented by graphs whose vertices are responsible for information processing. This family of approaches includes not only overly popular neural models, but also less popular models, in particular Hidden Markov Model (HMM) and Fuzzy Cognitive Map (FCM). In this paper, we present a systematic study on the application of HMMs and FCMs to time series classification. 

We emphasize that there are already two modes of using HMMs for pattern classification. The first, in which one HMM is constructed for each class. The second, in which one HMM is built for each time series in the dataset to be classified. Both variants are present in the literature on standard pattern classification, and our contribution is to apply these two models to time series classification. However, no such methodology exists for Fuzzy Cognitive Maps. Therefore, we transfer the methodology of pattern classification using HMM to the area of time series classification using FCMs. That is, we propose two variants of time series classification using FCM. The first, in which one FCM is built for each class. The second, in which one FCM is built for each time series. These two variants of using FCMs to  process data are new to the literature on time series and, in general, pattern classification.  

In an empirical study using a set of \varDataSetsNumber~benchmark datasets, we analyze and compare the effectiveness of the proposed approaches. 
Experiments showed that the one-model-per-time-series variant outperforms the one-model-per-class variant in terms of classification accuracy. In the case of HMMs, this variant, also requires less training time. In the case of FCMs, it is more time consuming than the one-model-per-class scheme.
The choice between HMM and FCM should depend on a dataset.

The paper is structured as follows. Section~\ref{sec:Literature Review} presents the basic knowledge of time series classification. Sections~\ref{sec:preliminaries} and \ref{sec:method} present preliminary background knowledge and the methods investigated. Section~\ref{sec:Experiment} is devoted to the experimental evaluation of the proposed approaches. Section~\ref{sec:Conclusion} concludes the paper.

\section{Literature Review on Time Series Classification}
\label{sec:Literature Review}

Time series classification is a~thriving area~of research.  We can distinguish six types of approaches in this field. 

The first group of approaches uses elastic measures of time series similarity, such as  Dynamic Time Warping (DTW).
Its purpose is to transform time series to facilitate their similarity assessment. 
DTW  finds the~so-called warping path, which ensures that the~distance between two sequences is the~smallest. There is an impressive amount of literature discussing different variants of DTW  applied to time series classification. 
For instance,  in~\cite{RakthanmanonCampanaMueenBatistaWestover2013} we find a~description of DTW with  performance enhancing corrections, while in~\cite{Jeong2011} an algorithm called WDTW -- Weighted DTW was introduced. 
A~similar idea~was conveyed under the~name Flexible DTW (FDTW) in \cite{Hsu2015}. FDTW  adds additional scoring to promote contiguously long one-to-one time series fragments. 

The~second group of methods uses the~concept of time series shapelets. A~shapelet is a~subsequence of a~given time series that represents well regularities in the~time series.
Methods based on shapelets are particularly suitable for datasets where there is a clear pattern in the values of the time series that uniquely identifies the class label, but the location of this pattern is not fixed. 
A~simple algorithm that can be applied in such a~case is the~shapelet-based method presented in~\cite{Ye2011}, which uses one shapelet per time series. 

The third group of methods is based on time series segmentation. Such methods usually segment the time series into intervals and then extract numerical features that are the basis for building a classifier. An example method implementing segmentation is Time Series Forest presented in~\cite{Deng2013}.

A~significant volume of research  has been put into the~development of dictionary-based algorithms. Two very popular methods of this type are Bag of Patterns (BOP) \cite{Lin2012} and Bag of Symbolic Fourier Approximation Symbols (BOSS) \cite{Schafer2015}. 
Large et al.~\cite{Large2019} improved BOSS by introducing orderless feature histograms to represent the~data. 

The fifth group of methods uses deep neural networks to perform the task of time series classification. This goup includes algorithms such as ROCKET \cite{Dempster2020}, TS-CHIEF \cite{Shifaz2020}, and InceptionTime \cite{Fawaz2020}.

The group of methods most relevant to this paper are model-based approaches. They are less popular than distance-based or subsequence-based methods. In this group are classifiers based on FCM and HMM. They are addressed in a~separate section of this paper.

To conclude the general literature review, it is worth mentioning that Hidden Markov Models or their extensions are used in the domain of time series classification typically not as main algorithms, but as supporting algorithms \cite{Antonucci2015,Bilal2012}. Fuzzy Cognitive Maps, on the other hand, are already present in the literature of time series classification \cite{Homenda2020}, but they are used in a different manner. In particular, trained FCM weight matrices are used to classify time series. In contrast, in this study we use FCM responses as a base for classification.

\section{Preliminaries} 
\label{sec:preliminaries}

\subsection{Introductory Notes on the~Hidden Markov Model}

Hidden Markov Model (HMM) is a~statistical method for modeling systems.  
It is a~Markov chain with hidden states.
A~Markov chain is a~stochastic process with the assumption that change depends only on the~current state.
In other words, the value assumed by a random variable $S$ at time $t$ depends only on the~value recorded at time $t-1$.
This assumption is called the~Markov property.
In the~Hidden Markov Model, we deal with a~compound Markov chain $(S_t, O_t)^\infty_{t=1}$, in which $(S_t)^\infty_{t=1}$ is a~Markov chain and $(O_t)^\infty_{t=1}$ is a~sequence of random variables.
The values assumed by the~random variable $S_t$, denoted as $s_t$, form the~state space of the HMM.

Let us focus on the~case, when $S_t$ is a~discrete random variable.
The values assumed by the~random variable $O_t$, denoted as $o_t$, are called emissions or observations. 
They can be either discrete or continuous.
If the observations are discrete, we usually generate them from a~categorical distribution.
If they are continuous, a Gaussian distribution can be used.
In an HMM, we observe only the values of the~variable $O_t$, while the~underlying values of $S_t$ remain hidden -- hence the~name ``hidden''. 
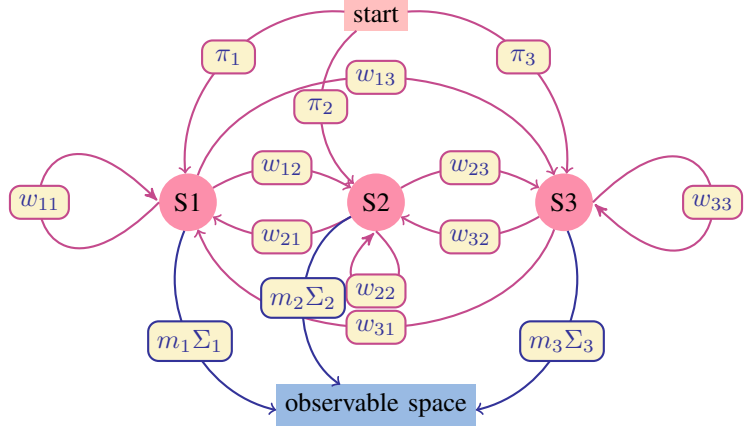
\begin{figure*}
	\centering
	\begin{tikzpicture}
		\SetGraphUnit{2.5}
		\tikzset{VertexStyle/.style = {shape = circle,fill = flamingopink}}
		\tikzset{EdgeStyle/.style={->,bend left =30, mulberry}}
		\Vertex{S2}
		\WE(S2){S1}
		\EA(S2){S3}
		\Edge[label = $w_{12}$](S1)(S2)
		\Edge[label = $w_{23}$](S2)(S3)
		\Edge[label = $w_{32}$](S3)(S2)
		\Edge[label = $w_{21}$](S2)(S1)
		\Loop[dist = 3cm, dir = NO, label = $w_{11}$](S1.west)
		\Loop[dist = 3cm, dir = SO, label = $w_{33}$](S3.east)
		\Loop[dist = 1.5cm, dir = WE, label = $w_{22}$](S2.south)
		\tikzset{EdgeStyle/.append style = {bend left = 70}}
		\Edge[label = $w_{13}$](S1)(S3)
		\Edge[label = $w_{31}$](S3)(S1)
		
		\tikzset{VertexStyle/.style = {shape = rectangle,yshift=-0.2cm,fill = carolinablue}}
		\SO(S2){observable space}
		
		\tikzset{EdgeStyle/.style={->,bend left =-50, blue(pigment)}}
		\Edge[label = $m_{1}\Sigma_1$](S1)(observable space)
		\tikzset{EdgeStyle/.style={->,bend left =-60, blue(pigment)}}
		\Edge[label = $m_{2}\Sigma_2$](S2)(observable space)
		\tikzset{EdgeStyle/.style={->,bend left =50, blue(pigment)}}
		\Edge[label = $m_{3}\Sigma_3$](S3)(observable space)
		
		\tikzset{VertexStyle/.style = {shape = rectangle,yshift=5.0cm,fill = pink}}
		\SO(S2){start}
		
		\tikzset{EdgeStyle/.style={->,bend left =-50, mulberry}}
		\Edge[label = $\pi_{1}$](start)(S1)
		\tikzset{EdgeStyle/.style={->,bend left =-50, mulberry}}
		\Edge[label = $\pi_{2}$](start)(S2)
		\tikzset{EdgeStyle/.style={->,bend left =50, mulberry}}
		\Edge[label = $\pi_{3}$](start)(S3)
	\end{tikzpicture}
	\caption{A~generic HMM with three states and Gaussian emissions.}
	\label{fig:hmm_draw}
\end{figure*}

Interestingly, there is an analogous computational model in the~literature called probabilistic automaton,~\cite{Rabin1963}. However, we will not pursue investigation in this direction since probabilistic automata are simply a generalization of the Markov chain model.

In order to learn the parameters of the HMM, a data-driven optimization procedure is necessary. 
Although there is no feasible exact algorithm to solve this problem, we have other methods available. Among several commonly used approaches is the Baum-Welch algorithm. It is based on the expectation-maximization (EM) algorithm,~\cite{Jelinek1975}. 
EM looks for the~maximum likelihood estimate of the~unknown parameters.
It is a~good choice when direct maximization of the~log-likelihood function is difficult.

\subsection{Introductory Notes on Fuzzy Cognitive Maps}

\begin{figure*}
	\centering
	\begin{tikzpicture}
		\SetGraphUnit{2.5}
		\tikzset{VertexStyle/.style = {shape = circle,fill = flamingopink}}
		\tikzset{EdgeStyle/.style={->,bend left =30, mulberry}}
		\Vertex{C2}
		\WE(C2){C1}
		\EA(C2){C3}
		\Edge[label = $w_{12}$](C1)(C2)
		\Edge[label = $w_{23}$](C2)(C3)
		\Edge[label = $w_{32}$](C3)(C2)
		\Edge[label = $w_{21}$](C2)(C1)
		\Loop[dist = 3cm, dir = NO, label = $w_{11}$](C1.west)
		\Loop[dist = 3cm, dir = SO, label = $w_{33}$](C3.east)
		\Loop[dist = 1.5cm, dir = WE, label = $w_{22}$](C2.south)
		\tikzset{EdgeStyle/.append style = {bend left = 80}}
		\Edge[label = $w_{13}$](C1)(C3)
		\Edge[label = $w_{31}$](C3)(C1)
	\end{tikzpicture}\vspace{-5pt}
	\caption{A~generic FCM with three concepts C1, C2, and C3.}
	\label{fig:fcm_draw}
\end{figure*}
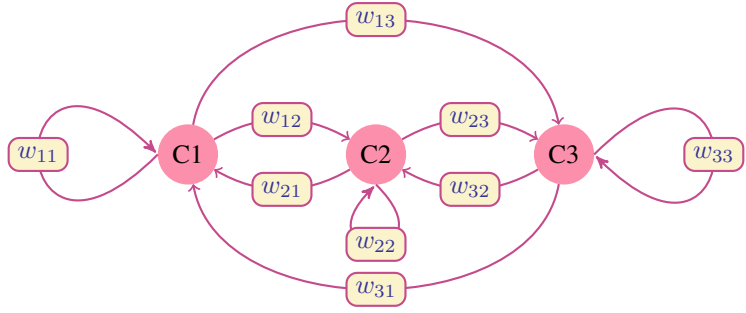

In the~focal point of this paper stand Fuzzy Cognitive Maps (FCMs), an information representation scheme designed to model temporal data.
FCMs were developed by B. Kosko in 1986 \cite{Kosko1986} as a~flexible extension of plain Cognitive Maps.
FCMs represent temporal phenomena~in the~form of a~weighted directed graph. A~sketch of a~generic FCM with three nodes is given in Fig.~\ref{fig:fcm_draw}. The nodes in an FCM are called concepts.

FCMs process data~according to the~following formula:
\begin{equation}
	\label{eq:reasoning}
	x_{i}^{(t+1)} = f \left(\sum_{j=1}^{P} w_{ji} x_{j}^{(t)} \right)
\end{equation}
$x_{i}^{(t+1)}$ is the~output of the~$i$th concept. It is computed for the~moment in time $(t+1)$ based on input that was observed at the~moment in time $t$.
Input and output, that is $a$, belongs to the~interval $[0,1]$.
$P$ is the~number of concepts in the~FCM.
$w$ denote weights, $w_{ij} \in [-1,1], i,j, =1, \ldots, P$.
$f$ is a~sigmoid function given in Eq.~(\ref{eq:sigmoid}):
\begin{equation}
	\label{eq:sigmoid}
	f(x)=\frac {1}{1+e^{-\tau \cdot x}}
\end{equation}
$x$ is in the~domain of real numbers from $(-\infty, \infty)$. $\tau$ is a~parameter controlling the~steepness of the~sigmoid function, which resembles the~S-shaped curve. Many papers on FCMs set $\tau$ to 2.5 or 5. $f$ squashes the~output to the~interval $[0,1]$.

It becomes apparent that the~FCM processing model, see Eq.~(\ref{eq:reasoning}), is suitable for describing the~behavior of dynamical systems and time series in general. Of course, due to the~finite nature of the~model it is of particularly useful for data~with seasonal and cyclic patterns.

FCM training boils down to the~task of finding the~values of weights between the~concepts. This is usually accomplished through an iterative procedure that adjusts the~weights to minimize the time series prediction one step forward. The~process uses historical time series data~and minimizes the Sum of Squared Errors (SSE) given in Eq.~(\ref{eq:sse}).	
\begin{equation}
	\label{eq:sse}
	SSE = \sum_{i=1}^N\sum_{j=1}^{P}(x_{ji}-y_{ji})^{2}
\end{equation}
$y_{ji}$ is the~predicted value of $x_{ji}$. $N$ is the~number of observations in the~set for which we calculate the~SSE.

Minimization of Eq.~(\ref{eq:sse}) is usually performed using heuristic searches \cite{Salmeron2019}. In this study, we use Differential Evolution. Thus, due to space limitations we restrict the discussion of the applied optimization method to the empirical study and skip theoretical explantation.

We use fuzzified time series as input data, which is used to create FCM model. The~fuzzification step is implemented using the fuzzy c-means algorithm. 
We launch the~clustering algorithm and obtain centroids. They are representing fuzzy sets and generalize the~underlying time series values. 
We run the clustering procedure for a data set with a two-dimensional representation of the time series  $\big( (z_2, dz_2), (z_3, dz_3) \ldots, (z_N, dz_N) \big)$, where $dz_i = z_i - z_{i-1}$ and $z_i$ is an $i$th element of the~scalar time series.	

We can link each original time series data~point (a~pair, $\boldsymbol{z_i} = (z_i, dz_i)$) with each centroid using the~following formula:
\begin{equation}
	x_{ij} = \frac{1}{\sum\limits_{k=1}^P \Big( \frac{\Vert \boldsymbol{z}_i - \boldsymbol{v}_j \Vert}{\Vert \boldsymbol{z}_i - \boldsymbol{v}_k \Vert} \Big)^{2/(M-1)} }
	\label{eq:fcmeans_memfun}
\end{equation}
where $\boldsymbol{v}$ denotes a~centroid and $M$ is a~fuzzification coefficient. Centroids become FCM concepts.	
The~membership formula~given in Eq.~(\ref{eq:fcmeans_memfun}) produces data~that is ready to be used with an FCM that we will use to classify time series. Processing data~on the~level of fuzzy memberships to concepts C1, C2, ... does not impose any limitations, because our ultimate goal is to construct a~classifier.

\section{Time Series Classification with Fuzzy Cognitive Maps and Hidden Markov Model}
\label{sec:method}

\subsection{Time Series Classification with Hidden Markov Model}
\label{TimeSeriesClassificationWithHMM}

The~standard approach consists of training one Hidden Markov Model for each class (using all its time series) and classifying sequences to the~class which model yields the~highest probability of generating the~sequence. This strategy will be denoted as HMM~1C. It is also possible to create one model for each training sequence. Then, we can construct a~classifier of our choice using probabilities returned by all these models. We use Nearest Neighbour as the~classification rule and refer to this method as HMM~NN. Please note that the~numbers of parameters in these two approaches are very different, as the~number of trained models is much bigger when they correspond to individual sequences.

When continuous observations are concerned, emission probability of a~hidden state is usually expressed in the~form of Gaussian distribution or Gaussian Mixture distribution. Usage of other distributions (e.g. Poisson) is also possible, but less common. As we want to limit the~number of trainable parameters to be able to learn on very short sequences, we use a~single Gaussian distribution. The~likelihood function maximized with the~Baum-Welch algorithm will be optimized over three sets of parameters: probability of starting in each state, probabilities of transitioning to other states, and parameters governing the~emission functions, in our case defined by means and covariance matrices. Since it is possible (and often advisable) to put constrains on the~covariance matrices, we examine models with three types of matrices: spherical, diagonal and unconstrained. 
Baum-Welch algorithm is susceptible to getting stuck in local optima. To deal with this problem at least partially, we perform a few learning attempts from random starting points and choose the~one that gives the~highest value of the~log-likelihood function.

\subsection{The~Contribution -- a~Novel Approach to Time Series Classification with Fuzzy Cognitive Maps}

Our approach to time series modeling with Fuzzy Cognitive Maps is almost a~mirror image of the~two techniques presented in \ref{TimeSeriesClassificationWithHMM}, but with Hidden Markov Models swapped for Fuzzy Cognitive Maps. Every series (or class) is modeled as Fuzzy Cognitive Map learned with Differential Evolution to minimize the~Mean Squared Error between the~map's output and the~real value for every pair of consecutive observations. When a~new series is classified, the~class of the~model giving the~lowest MSE is chosen. Just like with HMM, we examine both one-model-per-class (FCM~1C) and one-model-per-series (FCM~NN) approaches, using Nearest Neighbour classification rule in the~latter.

\section{Experimental Evaluation}
\label{sec:Experiment}

\subsection{Data~Sets and Experimental Setup}

In this section, we address with the conducted empirical experiments. These involved \varDataSetsNumber~datasets. All datasets come from publicly available repository  \url{http://www.timeseriesclassification.com}.
The experiments performed made the~following common assumptions:
\begin{itemize}
	\item The~code was implemented in Python 3.7. We used the~following external Python libraries: numpy, pandas, scipy, scikit-learn, hmmlearn, sktime, tqdm, pathlib.
	\item The~repository we used provides a~ready division between training and test sets. The training sets were used for training purposes, but we split them into train and validation sets to perform 3-fold cross-validation.
	\item There were times when the~Hidden Markov Models  training procedure failed to create a valid model.  This happened almost exclusively when unconstrained covariance matrices in one-model-per-sequence approach were used. When such a~failure occurred, the~accuracy of the~method was set to zero for that cross-validation fold. This is an example of a~rigorous approach -- one could simply ignore the~faulty model (especially when it corresponds to only one training sequence) and perform classification using all the~remaining models. However, the~occurrence of a~failure would still suggest that the~approach used is not well-suited to the~problem.
\end{itemize}
The experiments were divided into stages. In the first stage, hyperparameters were tested for the FCM~NN and FCM~1C models. Several combinations of DE hyperparameters used for FCM optimization were tested. We do not include the results here due to space limitations. One of universally optimal set of parameters was found to be as follows: maximum number of iterations  $150$, mutation $0.5$, recombination $0.5$, and popsize $10$. The only hyperparameter that needs to be tuned individually for each dataset is the number of concepts forming a map.

The hyperparameters for the HMM~1C and HMM~NN optimization were tested similarly. The outcomes of these tests show that we can set the maximum number of  iterations performed to $50$ and random initializations to $10$ for all experiments. The issue of how the algorithm behaves for different covariance matrices proved to be more challenging and we decided to make it a hyperparameter to be tuned individually for each dataset. Also, the number of hidden states was adjusted individually based on the accuracy of cross-validation.	

\subsection{Achieved Results}

\begin{figure}
	\centering
		\centering
		\includegraphics[width=0.36\textwidth]{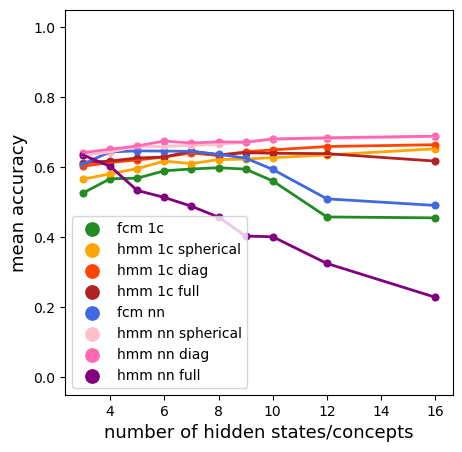}
		\caption{Average accuracy for all \varDataSetsNumber~datasets achieved with studied models.}
		\label{fig:AllDatasets}
\end{figure}

Fig.~\ref{fig:AllDatasets} shows the average classification accuracy on the~whole cross-validated training data. We observe a systematic superiority of the NN models. The only exception is HMM~NN with unconstrained covariance matrices that was overall the~worst of all tested methods. 

\begin{figure}
		\centering
		\includegraphics[width=0.36\textwidth]{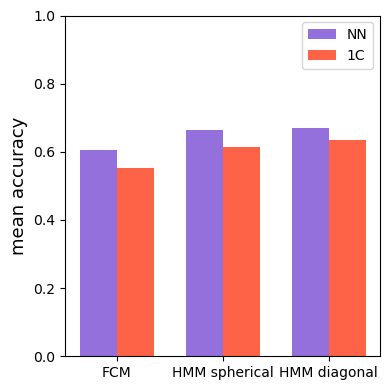}
		\caption{Clear advantage of NN methods over 1C methods.}
		\label{fig:1NN1C}
\end{figure}

Fig.~\ref{fig:1NN1C} compares models in which either FCM or HMM was built separately for each time series (the NN models) with models where either FCM or HMM was built one in each class (the 1C models). We skipped HMM~NN with unconstrained covariance matrices which provided unfeasible results and, in addition, posed challenges at the optimization process. The plot shows clear advantage of the NN models.

\begin{figure}
	\centering
	\includegraphics[width=0.36\textwidth]{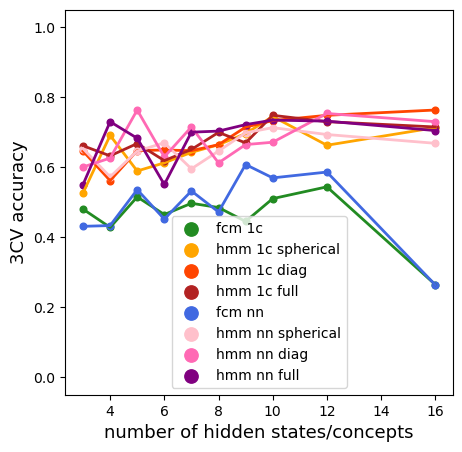}
	\caption{Advantage of HMM methods. The case of Car dataset.}
	\label{fig:Car}
\end{figure}

\begin{figure}
	\centering
	\includegraphics[width=0.36\textwidth]{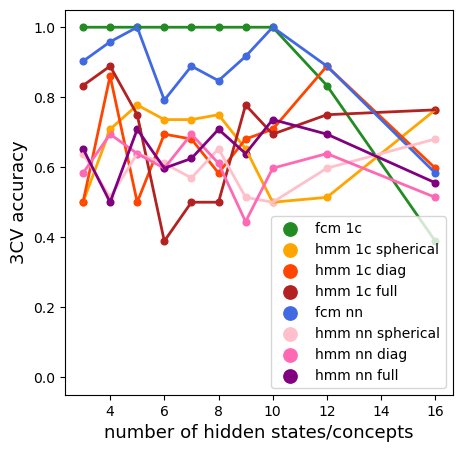}
	\caption{Advantage of FCM methods. The case of ShapeletSim dataset.}
	\label{fig:ShapeletSim}
\end{figure}

The comparison of different models revealed that there is no one clear winner, the performance depends on the dataset. In  Fig.~\ref{fig:ShapeletSim}, and Fig.~\ref{fig:Car}, we show cases when different models provide systematically better results.

It is possible to highlight groups of methods that produced highly correlated results (measured with Spearman's Correlation). Results of HMM~1C with different covariance matrices were almost interchangeable (SC~$\approx0.97$). The same was true for HMM NN spherical coupled with HMM NN diagonal. Similarities (FCM~1C $\approx$ FCM~NN) and (HMM~1C $\approx$ HMM~NN) were not that high (SC~$\approx0.85$). Correlations among 1C methods and NN methods were even lower (SC~$\approx0.8$), but not as low as between models that did not share any common trait (SC~$\approx0.72$). 

\begin{figure}
	\centering
	\includegraphics[width=0.36\textwidth]{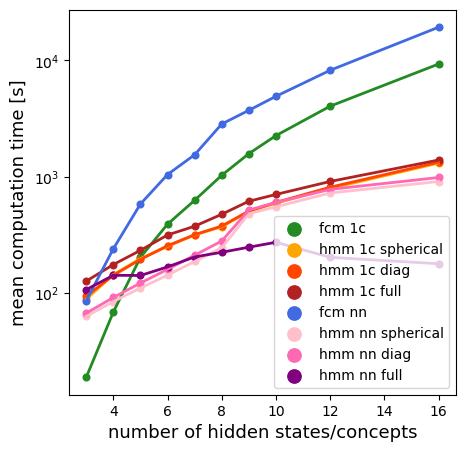}
	\includegraphics[width=0.36\textwidth]{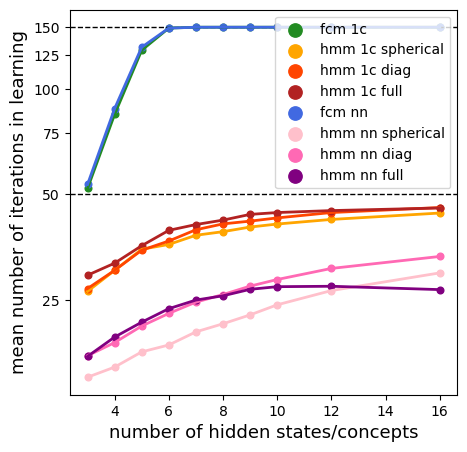}
	\caption{Average time of execution and average number of iterations while learning a~single model for all tested datasets. Please note the~logarithmic scale.}
	\label{fig:times}
\end{figure}

Despite similar accuracy results, the~execution time of the methods differed significantly. As shown in Fig.~\ref{fig:times}, the execution time of  Differential Evolution increased much faster than that for the Baum-Welch method even after the~maximum number of iterations in DE was reached. Execution time per iteration in both DE and Baum-Welch depends linearly on the~number of observations, regardless of whether learning is performed on all observations in the class or on each sequence separately. Thus, the only difference in computation time between NN and 1C methods stems from different behaviour of the optimizers. The average execution time of the HMM~NN method was lower than for the HMM~1C, while the average execution time of the FCM~NN method was higher than for the FCM~1C.

Subsequently, we present classification accuracy provided by the best model in each category (FCM~1C, FCM~NN, HMM~1C, and HMM~NN) trained for each dataset. Table~\ref{tab:results_variant2} presents this comparison. It also contains information about winning model parameters. All accuracies concern test sets. All experiments were conducted with 3-fold cross-validation. The table contains also the number of classes in each dataset to allow establishing a relative quality of the results.

\begin{table*}[!ht]
	\renewcommand{\arraystretch}{1.25}
	\caption{Classification accuracy (in \%) provided by the four studied models together with winning hyperparameters. Best result for each dataset is in bold.  In the case of HMMs there are two hyperparameters:  the number of states and covariance matrix type. In the case of FCMs, there is one hyperparameter -- nodes number.}
	\label{tab:results_variant2}
	\centering
	\scriptsize{
		\begin{tabular}{@{}r|r|r|r|r|r|r|r|r|r@{}}
			\hline
			\hline
			\multirow{3}{*}{data~set} &\multicolumn{1}{c|}{\# of}&\multicolumn{4}{|c}{HMM}&\multicolumn{4}{|c}{FCM}\\ 
			\cline{3-10}
			& \multirow{2}{*}{classes}&\multicolumn{2}{c|}{HMM~1C} & \multicolumn{2}{c|}{HMM~NN}&\multicolumn{2}{c|}{FCM~1C} & \multicolumn{2}{c}{FCM~NN}\\ 
			\cline{3-10}
			& &accuracy&hpar&accuracy&hpar&accuracy&hpar&accuracy&hpar\\
			\hline
			\hline
			Adiac	&37&	58.82	&	4 full	&	60.61	&	8 full	&	66.50	&	6	&	\textbf{71.10}	&	6	\\\hline
			ArrowHead	&3&	53.14	&	16 full	&	\textbf{65.71}	&	5 full	&	44.57	&	10	&	62.86	&	3	\\\hline
			Beef	&	5&40.00	&	12 full	&	53.33	&	9 diag	&	43.33	&	4	&	\textbf{56.67}	&	5	\\\hline
			BeetleFly	&2&	\textbf{85.00}	&	3 diag	&	75.00	&	3 diag	&	55.00	&	4	&	75.00	&	5	\\\hline
			BirdChicken	&2&	\textbf{90.00}	&	6 sphe	&	\textbf{90.00}	&	6 full	&	75.00	&	16	&	60.00	&	16	\\\hline
			Car	&	4&\textbf{70.00}	&	16 diag	&	58.33	&	5 diag	&	53.33	&	12	&	58.33	&	9	\\\hline
			CBF	&	3&93.89	&	4 sphe	&	\textbf{97.67}	&	4 sphe	&	62.33	&	6	&	66.22	&	3	\\\hline
			Computers	&2&	62.40	&	6 diag	&	\textbf{74.00}	&	3 full	&	68.00	&	9	&	67.60	&	5	\\\hline
			Dist.PhalanxOut.AgeGroup&3	&	\textbf{72.10}	&	8 sphe	&	70.65	&	4 diag	&	43.12	&	12	&	71.38	&	4	\\\hline
			DistalPhalanxTW	&6&	63.31	&	16 full	&	61.15	&	4 sphe	&	\textbf{64.03}&	8	&	56.83	&	5	\\\hline
			ECG200	&2&	71.00	&	16 diag	&	79.00	&	10 sphe	&	53.00	&	16	&	\textbf{81.00}	&	5	\\\hline
			ECG5000	&5&	91.71	&	16 full	&	\textbf{92.49}	&	5 sphe	&	85.07	&	9	&	89.76	&	4	\\\hline
			ECGFiveDays	&2&	55.17	&	6 full	&	\textbf{82.23}	&	10 sphe	&	75.73	&	9	&	67.83	&	8	\\\hline
			FaceAll	&14	&63.85	&	12 diag	&	\textbf{78.28}	&	16 sphe	&	45.74	&	8	&	48.52	&	8	\\\hline
			FaceFour	&4&	69.32	&	16 diag	&	62.50	&	9 sphe	&	\textbf{68.18}	&	10	&	55.68	&	4	\\\hline
			FacesUCR	&14&	67.12	&	16 diag	&	\textbf{73.66}	&	16 sphe	&	50.54	&	10	&	57.61	&	7	\\\hline
			Fish	&	7&50.86	&	6 sphe	&	47.43	&	4 full	&	55.43	&	5	&	\textbf{62.29}	&	5	\\\hline
			GunPoint&2	&	78.00	&	5 diag	&	\textbf{92.67}	&	10 diag	&	82.67	&	7	&	77.33	&	9	\\\hline
			Ham	&2&	\textbf{64.76}	&	3 sphe	&	48.57	&	4 full	&	62.86	&	10	&	58.10	&	4	\\\hline
			Haptics	&5&	28.90	&	8 full	&	26.62	&	16 sphe	&	\textbf{35.06}	&	6	&	25.97	&	10	\\\hline
			Herring	&2&	\textbf{67.19}	&	16 diag	&	56.25	&	16 diag	&	53.12	&	3	&	53.12	&	8	\\\hline
			InlineSkate&7	&	34.18	&	16 sphe	&	33.64	&	7 full	&	27.27	&	10	&	\textbf{38.36}	&	7	\\\hline
			InsectWingbeatSound&10	&	16.97	&	8 sphe	&	18.38	&	16 full	&	\textbf{23.79}	&	4	&	15.91	&	5	\\\hline
			ItalyPowerDemand	&	2&86.01	&	16 diag	&	\textbf{88.24}	&	7 diag	&	78.52	&	4	&	81.44	&	8	\\\hline
			Lightning2	&2&	63.93	&	7 full	&	\textbf{75.41}	&	12 diag	&	54.10	&	9	&	68.85	&	4	\\\hline
			Lightning7	&7&	58.90	&	4 full	&\textbf{60.27}	&	10 diag	&	47.95	&	6	&	42.47	&	4	\\\hline
			Mallat	&8&	79.91	&	5 diag	&	74.58	&	16 sphe	&	\textbf{88.02}	&	9	&	80.51	&	8	\\\hline
			Meat	&3&	80.00	&	16 diag	&	88.33	&	9 full	&	\textbf{93.33}	&	4	&	68.33	&	3	\\\hline
			MiddlePhalanxOut.Correct	&	2&\textbf{62.34}	&	3 full	&	50.00	&	9 sphe	&	59.74	&	7	&	54.55	&	8	\\\hline
			Mid.PhalanxOut.AgeGroup	&3&	71.13	&	16 diag	&\textbf{71.82}	&	16 diag	&	48.80	&	9	&	65.64	&	5	\\\hline
			MiddlePhalanxTW	&6&	50.65	&	10 diag	&	42.86	&	4 diag	&	\textbf{55.84}	&	5	&	47.40	&	3	\\\hline
			MoteStrain	&2&	\textbf{83.31}	&	8 diag	&	79.63	&	10 sphe	&	81.39	&	6	&	80.35	&	6	\\\hline
			OliveOil	&	4&40.00	&	6 sphe	&	40.00	&	6 full	&	\textbf{80.00}	&	7	&	76.67	&	7	\\\hline
			OSULeaf	&	6&64.88	&	16 sphe	&	77.69	&	10 sphe	&	\textbf{80.99}	&	10	&	76.03	&	5	\\\hline
			PhalangesOutlinesCorrect&2	&	63.75	&	16 sphe	&	70.05	&	10 sphe	&	56.53	&	12	&	\textbf{71.91}	&	5	\\\hline
			Plane&7	&	95.24	&	3 full	&	\textbf{100}	&	3 full	&	\textbf{100}	&	7	&	\textbf{100}	&	7	\\\hline
			Prox.PhalanxOut.AgeGroup&3	&	82.93	&	4 diag	&	82.93	&	4 diag	&	\textbf{83.90}	&	7	&	79.02	&	7	\\\hline
			ProximalPhalanxTW	&6&	62.89	&	3 diag	&	77.32	&	6 diag	&	66.67	&	3	&	\textbf{78.01}	&	4	\\\hline
			ShapeletSim	&	2&86.67	&	12 diag	&	67.22	&	10 full	&	\textbf{90.00}	&	3	&	\textbf{90.00}	&	5	\\\hline
			SonyAIBORobotSurface1	&	2&\textbf{95.51}	&	16 diag	&	92.01	&	8 full	&	86.19	&	7	&	82.70	&	3	\\\hline
			SonyAIBORobotSurface2	&2&	\textbf{88.67}	&	9 sphe	&	87.20	&	8 diag	&	77.33	&	8	&	78.38	&	8	\\\hline
			Strawberry	&	2&73.78	&	16 diag	&	93.24	&	10 full	&	71.35	&	7	&	\textbf{95.14}	&	8	\\\hline
			SwedishLeaf	&	15&\textbf{84.00}	&	16 sphe	&	85.92	&	6 full	&	73.76	&	9	&	78.88	&	9	\\\hline
			Symbols	&	6&\textbf{82.41}	&	5 diag	&	81.11	&	6 full	&	68.84	&	7	&	66.83	&	6	\\\hline
			SyntheticControl&6	&	94.33	&	16 sphe	&	\textbf{94.67}	&	9 sphe	&	66.33	&	9	&	66.67	&	5	\\\hline
			ToeSegmentation1&2	&	78.07	&	4 diag	&	80.26	&	3 sphe	&	76.32	&	5	&	\textbf{82.46}	&	6	\\\hline
			ToeSegmentation2	&2&	70.00	&	12 diag	&	\textbf{76.15}	&	10 sphe	&	69.23	&	7	&	71.54	&	7	\\\hline
			Trace&4	&	87.00	&	5 diag	&	\textbf{99.00}	&	3 sphe	&	93.00	&	3	&	97.00	&	3	\\\hline
			TwoLeadECG	&2&	\textbf{96.66}	&	10 full	&	94.03	&	9 diag	&	91.66	&	5	&	79.98	&	12	\\\hline
			Wafer&2	&	96.90	&	16 full	&	96.51	&	12 sphe	&	83.40	&	12	&	\textbf{98.26}	&	6	\\			
			\hline				\hline
		\end{tabular}
	}
\end{table*}

In about half of the studied datasets, FCM provided best models. In the other half, HMM did. What is interesting, the best results achieved using FCM are based on maps with less concepts than hidden states in a corresponding HMM. This is an important result speaking for the FCM model. 

\section{Conclusion and Critical Discussion}
\label{sec:Conclusion}
The paper presented an empirical study that aimed at the comparison of the effectiveness of Hidden Markov Model and Fuzzy Cognitive Map in the task of time series classification. Two schemes of data classification are compared for both models.  The first one requires training of an FCM or HMM for each time series in a dataset. The second requires training one FCM or HMM per class. The study demonstrated that the first variant provides higher classification accuracy both for FCM and HMM. What is more, it turned out to be computationally less expensive.  The second important development addressed in this paper was the transfer of pattern classification methodology from HMMs to FCMs. Experiments show that FCMs used in this manner achieve similar classification accuracy as HMMs for similar model set-ups. The downside of the FCM~NN variant is its training time, which is higher than FCM~1C and that of HMMs.

Last but not least, we shall underline that the discussed methodology of time series classification using FCMs is not the only one present in the literature. There exist an FCM-based methodology introduced by Homenda and Jastrzebska \cite{Homenda2020}, which, all in all, achieves higher classification accuracies than the method presented in this paper. Nonetheless, we believe that there is a value added in studies on alternative, distinct methods.

\bibliographystyle{IEEEtran}  
\bibliography{mybibfile}

\end{document}